\documentclass[lettersize,journal]{IEEEtran}
\usepackage{amsmath,amsfonts}
\usepackage{algorithm}
\usepackage[noend]{algpseudocode}
\usepackage{array}
\usepackage{hyperref}
\usepackage{multirow}
\usepackage[caption=false,font=normalsize,labelfont=sf,textfont=sf]{subfig}
\usepackage{textcomp}
\usepackage{stfloats}
\usepackage{url}
\usepackage{verbatim}
\usepackage{graphicx}
\usepackage{cite}
\usepackage{subcaption}

\usepackage{color}
\begin{document}
\title{Transformer-Driven Multimodal Fusion for Explainable Suspiciousness Estimation in Visual Surveillance}
\author{Kuldeep Singh Yadav,~\IEEEmembership{Member,~IEEE} and Lalan Kumar,~\IEEEmembership{Member,~IEEE}
\thanks{Kuldeep Singh Yadav is with the Big Data Research and Supercomputing Division, CSIR Fourth Paradigm Institute, Bengaluru, India. (e-mail: kuldeep.4pi@csir.res.in).}%
\thanks{Lalan Kumar is with the Department of Electrical Engineering, Bharti School of Telecommunication, Yardi School of Artificial Intelligence, IIT Delhi, New Delhi, India (e-mail: lkumar@ee.iitd.ac.in).}
\thanks{Manuscript received xx, 2025}}


\maketitle
\begin{abstract}
Suspiciousness estimation is critical for proactive threat detection and ensuring public safety in complex environments. This work introduces a large-scale annotated dataset, \textbf{USE50k}, along with a computationally efficient vision-based framework for real-time suspiciousness analysis. The \textbf{USE50k} dataset contains 65,500 images captured from diverse and uncontrolled environments, such as airports, railway stations, restaurants, parks, and other public areas, covering a broad spectrum of cues including weapons, fire, crowd density, abnormal facial expressions, and unusual body postures.
Building on this dataset, we present \textbf{DeepUSEvision}, a lightweight and modular system integrating three key components, i.e., a Suspicious Object Detector based on an enhanced YOLOv12 architecture, dual Deep Convolutional Neural Networks (DCNN-I and DCNN-II) for facial expression and body-language recognition using image and landmark features, and a transformer-based Discriminator Network that adaptively fuses multimodal outputs to yield an interpretable suspiciousness score.
Extensive experiments confirm the superior accuracy, robustness, and interpretability of the proposed framework compared to state-of-the-art approaches. Collectively, the \textbf{USE50k} dataset and the \textbf{DeepUSEvision} framework establish a strong and scalable foundation for intelligent surveillance and real-time risk assessment in safety-critical applications.
\end{abstract}


\section{Introduction}
\label{sec:intro}
With rising global security threats, critical public infrastructures such as airports, banks, hospitals, and railway stations increasingly rely on large-scale video surveillance systems. These systems continuously capture visual data to detect unusual activities, including theft, unauthorized access, criminal identification, and anomaly detection through person tracking and behavior monitoring. Beyond merely tracking individuals, the ability to accurately identify potentially threatening objects and interpret the entire scene is crucial for proactive vigilance. Such capabilities are essential for recognizing a wide spectrum of suspicious events, ranging from loitering, unauthorized actions, and crowd mismanagement to high-risk incidents such as terrorist attacks, kidnappings, fires, and road accidents, thereby ensuring public safety and situational awareness \cite{salem2021surveillance,chiu2017multimodal,chen2020social}.
The massive volume of visual data generated by modern surveillance systems renders manual monitoring highly impractical and error-prone. Computer vision, a key domain of artificial intelligence, enables automated analysis and interpretation of such visual streams with high efficiency. By extracting high-dimensional representations from digital images and videos, computer vision facilitates intelligent decision-making for secure and real-time monitoring. Extensive research has demonstrated its effectiveness across diverse applications, including human activity recognition (HAR), image and video captioning, fall detection, audio-based activity analysis, traffic monitoring, object localization, region segmentation, and data classification \cite{HAR_2023_TPAMI,AR_2020_CVPR,image_caption,FALL_2017_TMC,audio_based,Traffick_management,seg_kd}.

Despite significant progress in computer vision, relatively limited research has focused specifically on Suspicious Activity Recognition (SAR) \cite{sus_Act,SAR_2007_TIP,SAR_2016_TKDE,SAR_2017_Access}, crime detection, and early threat prediction \cite{crime_det,Early_Threat_Detection}. Classical deep learning approaches such as Convolutional Neural Networks (CNN) and Long Short-Term Memory (LSTM) networks offer improved representational power, they are highly data-driven and require large-scale annotated datasets to achieve robust performance, a major limitation in surveillance scenarios where suspicious events are rare, diverse, and difficult to annotate. However, real-world surveillance presents several challenges that existing SAR and threat detection methods fail to address effectively. In addition, suspicious events are highly context-dependent, visually subtle, and occur infrequently, making it extremely difficult to construct large-scale, well-annotated datasets. 
Most recent approaches treat object detection, facial analysis, and behavioral understanding as isolated tasks, lacking a unified multimodal reasoning framework that can jointly interpret objects, emotions, and human intent within a scene. This leads to poor generalization under unconstrained conditions involving occlusions, crowd density variations, illumination changes, and complex human interactions. Furthermore, the absence of standardized, publicly available datasets reflecting realistic security, critical environments, restricts progress toward reliable and deployable SAR models.

To address these limitations, we propose a comprehensive surveillance framework that integrates a novel dataset, USE50k, and a robust multimodal threat assessment architecture, DeepUSEvision, for real-time suspiciousness estimation in uncontrolled environments. 
Building upon the novel dataset, DeepUSEvision employs a cascaded two-stage inference pipeline. By jointly integrating purpose-built data and a hierarchically cascaded decision model, the proposed system delivers a scalable, context-aware, and operationally viable solution for next-generation security surveillance.

\section{Related Work}
\label{sec:related}

Research in intelligent surveillance spans anomaly detection, real-time object perception, multimodal fusion, and explainable AI. Early and recent work in video anomaly and violence detection has relied on weakly supervised MIL frameworks for untrimmed surveillance footage (e.g., UCF–Crime), establishing large-scale benchmarks for real-world abnormality modeling \cite{Sultani2018}. Multimodal extensions, such as audio–visual violence detection (e.g., XD-Violence), further highlight the benefit of combining heterogeneous cues under noisy surveillance conditions \cite{Wu2020_XD}. Surveys such as \cite{Ullah2023} emphasize the need for richer annotations and interpretable mechanisms in safety-critical surveillance applications. However, existing datasets rarely provide frame-level joint annotation of objects, facial affect, and body language—an essential requirement for suspiciousness estimation.

Real-time detection is dominated by single-stage architectures, with the YOLO family pioneering fast, anchor-free detection for surveillance-scale deployments \cite{Redmon2016,Bochkovskiy2020}. Lightweight backbones such as MobileNetV3 offer favorable latency–accuracy tradeoffs and are widely adopted for constrained tasks like face, emotion, or pose analysis \cite{Howard2019_MobileNetV3}. These advances motivate adopting an optimized detector capable of identifying security-critical entities (persons, faces, weapons, fire) under unconstrained conditions while maintaining real-time throughput on edge and workstation hardware.

Recent transformer-based architectures have demonstrated strong capability for cross-modal fusion and long-range dependency modeling. Works such as TransFuser and multimodal attention transformers illustrate how self-attention can fuse heterogeneous perceptual inputs to support regression, control, or decision-making tasks \cite{Prakash2021_TransFuser,Sun2021_MultiModal}. Unlike classification-oriented fusion systems, suspiciousness estimation demands continuous scoring and fine-grained reasoning across cues such as object presence, emotional state, and body posture. The proposed transformer-based Discriminator addresses this by learning higher-order cue interactions (e.g., weapon–emotion synergy), enabling more principled fusion than static or heuristic-weighted approaches.

Explainability in vision systems has progressed through model-agnostic attribution techniques, particularly SHAP, which provides consistent feature- and interaction-level explanations suited for forensic security tasks \cite{Lundberg2017_SHAP,Shape}. Prior works integrate attention with explanation, yet rarely analyze interaction manifolds or contextual dependencies driving risk scoring. Our work extends this direction by combining transformer attention with SHAP interaction maps and 3D risk manifolds to expose how multimodal cues jointly influence suspiciousness.

Overall, existing literature provides strong foundations in anomaly detection, object perception, fusion, and explainability. However, key gaps remain: (i) the absence of datasets with unified annotations for objects, emotions, and body language; (ii) limited modeling of inter-cue interactions for continuous risk estimation; and (iii) insufficient explainability tailored for security decision workflows. The present work addresses these challenges through the introduction of the USE50k dataset, a transformer-driven multimodal fusion framework for suspiciousness prediction, and a comprehensive explainability pipeline for interpreting cue synergy under realistic surveillance conditions.

\section{USE50k}
\subsection{Data Collection}
The Uncontrolled Suspiciousness Estimation (USE50k) dataset comprises 65,500 images collected from public social media sources and manually curated surveillance footage. It encapsulates a wide spectrum of security-critical visual cues, including persons, faces, weapons, and fire, enabling fine-grained analysis of threat-indicative objects. In addition, the dataset captures behavioral intelligence signals such as facial expressions (seven emotion classes) and body language (normal vs. abnormal), which are crucial for intent understanding beyond object-level perception. The images span diverse real-world environments, including airports, parks, restaurants, malls, streets, railway and bus stations, ensuring strong generalization capability under unconstrained conditions. A detailed statistical breakdown is provided in Table~\ref{tab:use_db_overview}.
\begin{table}[htbp]
\centering
\caption{Statistical Overview of the USE50k Dataset}
\label{tab:use_db_overview}
\small
\begin{tabular}{|p{1.6cm}|p{6.2cm}|}
\hline
\textbf{Attribute} & \textbf{Details} \\ \hline
\# Images & 65,500 (Train: 50,000; Test: 15,500) \\ \hline
Object instances & Face: 0.4M; Person: 0.6M; Weapon: 35k; Fire: 10k \\ \hline
Resolution range & $640\times480$, $720\times640$, $1280\times720$, $1920\times080$, $3840\times2160$ (mixed, unconstrained) \\ \hline
Scene \& environment & Indoor / Outdoor; Airports, Parks, Restaurants, Malls, Streets, Railway \& Bus Stations \\ \hline
Lighting \& visual variation & Day, Night, Dim, Extreme illumination, Shadows, Motion blur, Occlusion \\ \hline
Annotation types & Instance bounding boxes, Facial-expression labels (7 classes), Body-language state (Normal/Abnormal), Scene-level suspiciousness score \\ \hline
\end{tabular}
\end{table}

\subsection{Data Annotation Using Deep Automated Annotation Tool}
Annotating large-scale surveillance datasets is labor-intensive and prone to human fatigue and inconsistency. To overcome this, a Deep Automated Annotation (DAA) tool \cite{yadav2025dlaat} was employed, powered by a lightweight 8-layer deep learning backbone customized for USE50k. The model was first trained to detect critical security-relevant entities such as persons, faces, weapons, and fire, enabling rapid generation of bounding-box annotations with minimal human intervention.
For higher-level behavioral context, including body language state and scene-level suspiciousness score, a hybrid human-in-the-loop strategy was adopted. Two domain experts independently reviewed and labeled the behavioral attributes, followed by a final senior annotator who performed manual verification to resolve any ambiguity or bias. This combined automation–expert pipeline ensures high annotation accuracy, consistency, and scalability while significantly reducing annotation time and manual strain.
\subsection{Data Augmentation}
To enhance model generalization and mitigate overfitting, a diverse set of data augmentation strategies was applied during training. The augmentation pipeline included spatial transformations such as translation and scaling (within ±10 pixels along both axes), horizontal flipping, and noise perturbation using Gaussian noise with intensity levels of 0.01–0.03. These augmentations simulate realistic variations in object position, scale, viewpoint, and visual quality, thereby improving the model’s robustness to unconstrained real-world surveillance conditions.

\section{DeepUSEvision}
\begin{figure*}[t!]
\centering
\centerline{\includegraphics[width=1\linewidth, trim=200 350 250 300, clip, angle=0]{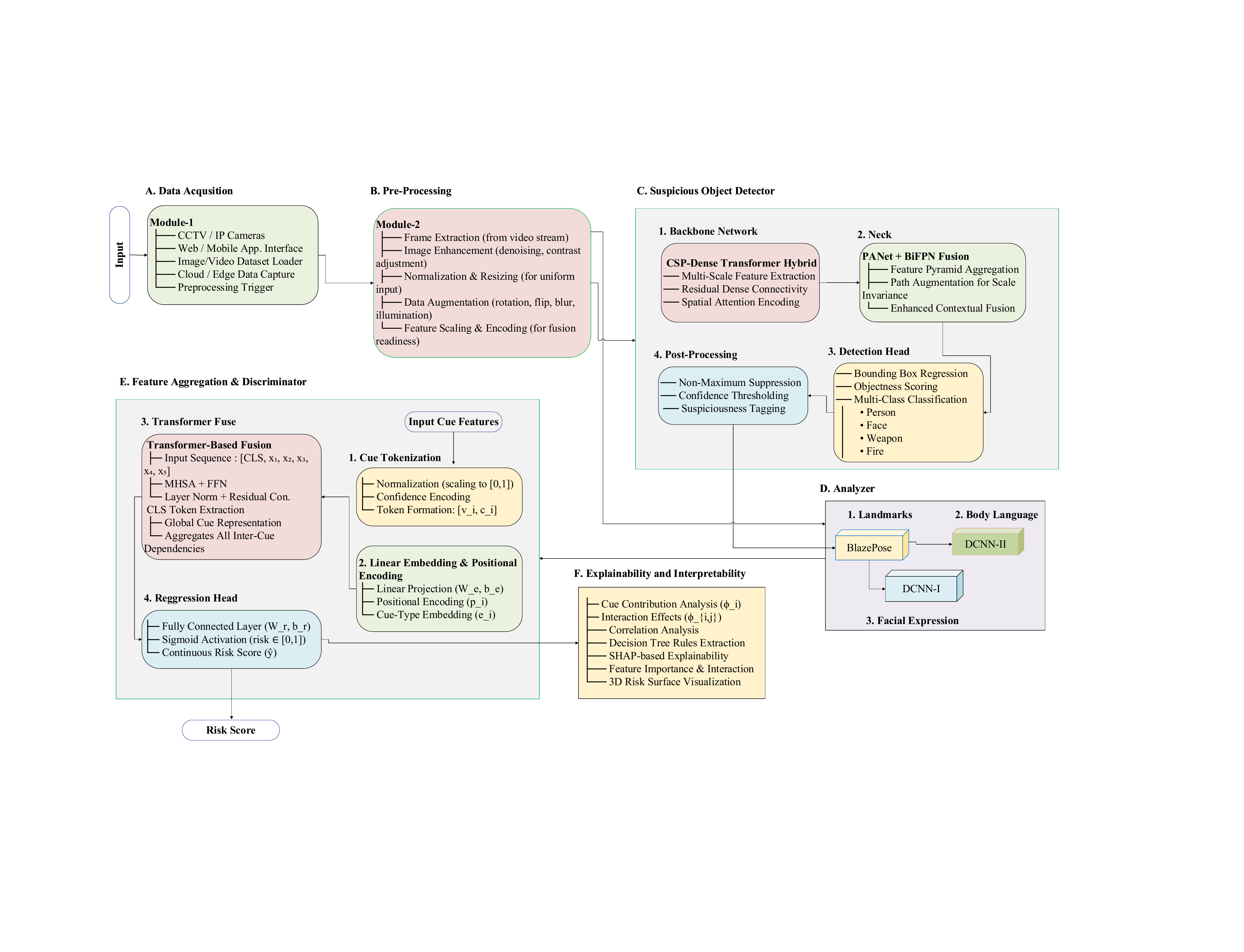}}
\caption{Block diagram of the proposed DeepUSEvision system}
\label{fig:fig1}
\end{figure*}

This section presents DeepUSEvision, an end-to-end deep learning framework for uncontrolled suspiciousness estimation in real-world surveillance environments. The system is composed of six key modules: (i) data acquisition, (ii) pre-processing, (iii) suspicious object detector, (iv) behavioral analyzer, (v) feature aggregation and transformer-based discriminator, and (vi) explainability and interpretability. The complete pipeline of the proposed framework is illustrated in Fig.~\ref{fig:fig1}.
The DeepUSEvision pipeline begins with real-time data acquisition from unconstrained surveillance feeds, typically captured under widely varying conditions such as indoor/outdoor scenes, day/night lighting, dense crowds, and partial occlusions. Each input frame is first standardized via a lightweight pre-processing routine consisting of contrast normalization, histogram equalization, denoising, and resolution scaling to preserve structural fidelity while suppressing illumination artifacts. This pre-conditioning ensures consistent visual quality and stabilizes downstream detection performance, especially under high-variance environments typical of real-world CCTV deployments.

\subsection{Suspicious Objects Detection}
\label{ssec:SOD}

The Suspicious Objects Detector (SOD) is the core perception module of the proposed DeepUSEvision system, responsible for detecting four high-risk cues, \textit{person, face, weapon, and fire}. These categories carry distinct semantic importance in suspiciousness estimation. The detector is built on a customized YOLOv12 backbone \cite{tian2025yolov12} that couples convolutional inductive priors with transformer-based global reasoning for robust discrimination under unconstrained surveillance conditions.

\textbf{Backbone (Hybrid Feature Extraction):}
The backbone integrates CSP-DarkNet with a Transformer Attention Module (TAM), combining local texture encoding with long-range dependency modeling. A modified C2f-Trans block introduces lightweight self-attention inside residual paths. Given an input map \( X \in \mathbb{R}^{C \times H \times W} \), the block output is:
\begin{equation}
Z = g(\text{concat}(f_{\text{conv}}(X_1), f_{\text{attn}}(X_2)))
\end{equation}
where \(X_1,X_2=\text{split}(X)\), \(f_{\text{conv}}\) captures local structures, and \(f_{\text{attn}}\) models global cues. This hybrid representation improves robustness to occlusion, illumination variation, and scale changes, common challenges in surveillance imagery.

\textbf{Neck (Multi-Scale Contextual Fusion):}
The neck uses an enhanced BiFPN with CSP-PAN for efficient top–down and bottom–up fusion. For feature levels \(F_3,F_4,F_5\), the fused multi-scale representation is:
\begin{equation}
F_{\text{fused}} = \sum_{i=3}^{5} \alpha_i \cdot \text{Align}(F_i)
\end{equation}
where learnable weights \(\alpha_i\) satisfy \(\sum_i \alpha_i=1\). This formulation emphasizes scale-adaptive information important for detecting small objects like weapons and large cues such as persons and fire.

\textbf{Detection Head (Decoupled Anchor-Free Inference):}
The detection head follows a decoupled, anchor-free design, predicting objectness, class scores, and bounding boxes independently for improved localization–classification disentanglement.  
The object-center probability is:
\begin{equation}
P_{\text{center}}=\sigma(\text{Conv}(F_{\text{BiFPN}})),
\end{equation}
while bounding box offsets are:
\begin{equation}
(b_x,b_y,b_w,b_h)=W_{\text{bbox}}F_{\text{BiFPN}} + b_{\text{bbox}},
\end{equation}
and class/confidence predictions are:
\begin{equation}
C=\text{Softmax}(W_{\text{cls}}F_{\text{BiFPN}}+b_{\text{cls}}), \quad
s=\sigma(W_{\text{conf}}F_{\text{BiFPN}}+b_{\text{conf}}).
\end{equation}

\textbf{Optimization and Training:}
The model is trained using a multi-task loss:
\begin{equation}
\mathcal{L}_{total}= \lambda_1\mathcal{L}_{\text{CIoU}} + \lambda_2\mathcal{L}_{\text{conf}} + \lambda_3\mathcal{L}_{\text{cls}},
\end{equation}
where CIoU enhances geometric consistency, BCE optimizes objectness, and focal loss handles class imbalance across the four suspicious object categories. Strong augmentations like brightness jittering, motion blur, and random occlusion, simulate real-world surveillance artifacts. 
Overall, the customized YOLOv12-based SOD provides high recall for critical cues and forms the foundation for downstream multimodal suspiciousness estimation.

\subsection{Facial Expression and Body Language Analyzer}
The behavioral analysis module jointly interprets facial expressions and body posture using 33 anatomical keypoints extracted via the BlazePose landmark detection framework~\cite{blazepose}. The first 10 landmarks correspond to facial regions, while the remaining 23 capture full-body skeletal dynamics. These landmarks are fused with pixel-space observations to form a compact, semantically rich input representation.
A lightweight MobileNetV3-inspired deep network is adopted as the unified backend for both tasks. The architecture consists of a stem convolution followed by five inverted residual blocks with progressive feature scales $(56 \times 56 \times 16 \rightarrow 28 \times 28 \times 24 \rightarrow 14 \times 14 \times 48 \rightarrow 7 \times 7 \times 96)$. Each block integrates depthwise separable convolutions, linear bottlenecks, and squeeze-and-excitation (SE) attention, enabling efficient channel recalibration while preserving computational efficiency.

This network is trained as DCNN-I for seven-category facial expression recognition (Anger, Disgust, Fear, Happy, Neutral, Sad, and Surprise), using a feature fusion of face crops and facial landmarks (0--10) to enhance emotion discriminability. It is independently trained as DCNN-II for binary body language analysis (Normal vs.\ Abnormal), using all 33 landmarks to model posture and behavioral cues. Both subnetworks are trained on facial images and suspicious data, respectively. This dual-stream behavioral inference provides highly complementary affective and kinematic cues, strengthening downstream suspiciousness estimation.

\subsection{Discriminator Network: Transformer-Based Fusion for Suspiciousness Estimation}
In this module, the \textit{Discriminator Network} for estimating the suspiciousness of a given scene based on multiple visual cues, including the number of persons ($\text{Nm\_P}$), suspicious emotions ($\text{Nm\_Em}$), weapons ($\text{Nm\_Wp}$), fire flags ($\text{Nm\_F}$), and abnormal body language ($\text{Nm\_BL}$) is proposed. Each cue is first detected using dedicated SOD modules, and then fused in a transformer-based architecture to generate a continuous \textit{risk score}.

Traditional regression or tree-based models treat each cue independently and often fail to capture complex interactions between cues. For instance, the simultaneous presence of a weapon and suspicious emotions may indicate a higher risk than each cue individually. To address this, our discriminator network models both \textit{individual cue contributions} and \textit{pairwise or higher-order interactions} using a transformer-based fusion mechanism.

The Feature Aggregation and Discriminator (FAD) module was implemented using a transformer-based fusion architecture designed to learn contextual dependencies among heterogeneous modalities. The fusion encoder consists of four transformer layers, each employing multi-head self-attention (8 heads) followed by layer normalization and feed-forward sublayers. The discriminator head comprises two fully connected layers with 128 and 64 neurons, respectively, activated using ReLU functions and batch normalization, and a final sigmoid neuron producing the continuous suspiciousness likelihood in the range $[0,1]$.

\subsubsection{Cue Tokenization and Representation}
Each detected cue is represented as a token vector:
\begin{equation}
\mathbf{t}_i = [v_i, c_i] \in \mathbb{R}^{2}, \quad i \in \{1,\dots,5\}
\end{equation}
where $v_i$ denotes the normalized cue value (e.g., number of persons scaled to $[0,1]$) and $c_i$ represents the confidence in detection. The set of cue tokens forms the input matrix:
\begin{equation}
\mathbf{T} = [\mathbf{t}_1, \mathbf{t}_2, \dots, \mathbf{t}_5] \in \mathbb{R}^{5 \times 2}.
\end{equation}

\subsubsection{Transformer-Based Fusion}
Each cue token is projected into a high-dimensional latent space using a learnable linear embedding:
\begin{equation}
\mathbf{x}_i = \mathbf{W}_e \mathbf{t}_i + \mathbf{b}_e, \quad \mathbf{x}_i \in \mathbb{R}^{d}
\end{equation}
where $d$ is the model dimension. Positional encodings and cue-type embeddings are added to incorporate sequence information:
\begin{equation}
\tilde{\mathbf{x}}_i = \mathbf{x}_i + \mathbf{p}_i + \mathbf{e}_i
\end{equation}
where $\mathbf{p}_i$ is the positional encoding and $\mathbf{e}_i$ is the cue-type embedding.

The augmented token sequence is prepended with a learnable classification token $\text{CLS}$:
\begin{equation}
\mathbf{X} = [\text{CLS}, \tilde{\mathbf{x}}_1, \tilde{\mathbf{x}}_2, \dots, \tilde{\mathbf{x}}_5] \in \mathbb{R}^{6 \times d}.
\end{equation}

The sequence is passed through a standard transformer encoder with multi-head self-attention (MHSA) layers to capture intra-cue and inter-cue interactions:
\begin{equation}
\mathbf{Z} = \text{TransformerEncoder}(\mathbf{X}) \in \mathbb{R}^{6 \times d}
\end{equation}

\subsubsection{Risk Score Estimation}
The output corresponding to the \text{CLS} token, $\mathbf{Z}_\text{CLS} \in \mathbb{R}^d$, serves as a fused representation of all cues. It is passed through a small feed-forward network to produce a continuous risk score:
\begin{equation}
\hat{y} = f_\text{reg}(\mathbf{Z}_\text{CLS}) \in \mathbb{R}
\end{equation}
where $f_\text{reg}$ is a learnable regression head, and $\hat{y}$ represents the estimated suspiciousness.

The multi-head self-attention mechanism inherently models pairwise and higher-order interactions between cues.
Multiple heads allow the network to learn complementary interactions simultaneously.

\subsection{Explainability and Interpretability Analysis}

To ensure transparency of the proposed \textit{DeepUSEvision} framework, an explainability pipeline was integrated into the Feature Aggregation and Discriminator module. The analysis combines correlation evaluation, surrogate modeling, feature importance assessment, and SHAP-based attributions to elucidate how multimodal cues influence the predicted suspiciousness score.

\textbf{Correlation and Dependency Analysis:}
Pearson and Spearman coefficients were computed to measure linear and monotonic dependencies. Both heatmaps indicated strong positive correlation between the risk score and $Nm_{Wp}$ and $Nm_{BL}$, highlighting their dominant influence in the overall inference.

\textbf{Surrogate Rule Extraction:}
A shallow Decision Tree Regressor (depth = 4) was trained as a surrogate to approximate the Transformer-based Discriminator. It provides interpretable hierarchical decision rules that approximate the black-box decision boundary while preserving major cue interactions.

\textbf{Global Feature Importance:}
Using mean decrease in impurity (MDI), the feature ranking was obtained as:  
$Nm_{Wp}$ (34.2\%), $Nm_{BL}$ (28.5\%), $Nm_{Em}$ (18.7\%), $Nm_{F}$ (11.4\%), and $Nm_{P}$ (7.2\%).  
This confirms that behavioral and contextual cues contribute more significantly to suspiciousness estimation than object-count cues alone.

\textbf{SHAP-Based Attribution:}
SHAP values were computed for unified global and local interpretability. The summary and dependence plots revealed that $Nm_{Wp}$ and $Nm_{BL}$ exert the strongest positive impact on suspiciousness, followed by $Nm_{Em}$ and $Nm_{F}$. SHAP interaction plots exposed non-linear relationships, particularly \textit{(Weapon–Body Language)} and \textit{(Emotion–Face)} pairings, validating the cross-modal coupling learned by the Transformer.

\textbf{Feature Interaction and Synergy:}
Pairwise SHAP interaction values were used to quantify joint cue effects. A Feature Synergy Network was constructed with node sizes proportional to feature importance and edges weighted by interaction strengths. The strongest interactions were: $(Nm_{Wp},Nm_{BL})~0.2971$, $(Nm_{Em},Nm_{F})~0.2146$, and $(Nm_{Wp},Nm_{Em})~0.1865$.  
3D response surfaces further illustrated the non-linear risk manifolds formed by these cue combinations, highlighting how perceptual and behavioral cues jointly modulate suspiciousness.

Overall, the explainability module confirms that the Discriminator effectively captures both main effects and higher-order cue synergies, yielding a transparent and interpretable suspiciousness estimation process.

\section{Results and Discussion}
\subsection{Experimental Setup}
The experiments are conducted on MATLAB and Python platforms, utilizing a robust system configuration comprising an Intel i7-13650HX processor, 64 GB RAM, and an NVIDIA Quadro P5000 GPU equipped with 16 GB of memory. This high-performance computational setup ensured the efficiency and reliability of extensive data analysis and comprehensive model evaluation.

\subsection{Suspicious Object Detection}
The Suspicious Object Detector (SOD) was trained on the USE50k dataset using a mini-batch size of~8 and 100~epochs (varied with parameters and optimized), where convergence was observed around the 90th epoch. A comprehensive augmentation pipeline, scaling, random cropping, rotation, color jittering, flipping, random erasing, and mosaic composition, was employed to simulate real surveillance conditions including occlusion, motion blur, illumination variation, and clutter.

\begin{figure*}[h]
    \centering
    \subfloat[]{
        \includegraphics[width=0.45\linewidth, trim=50 200 80 230, clip]{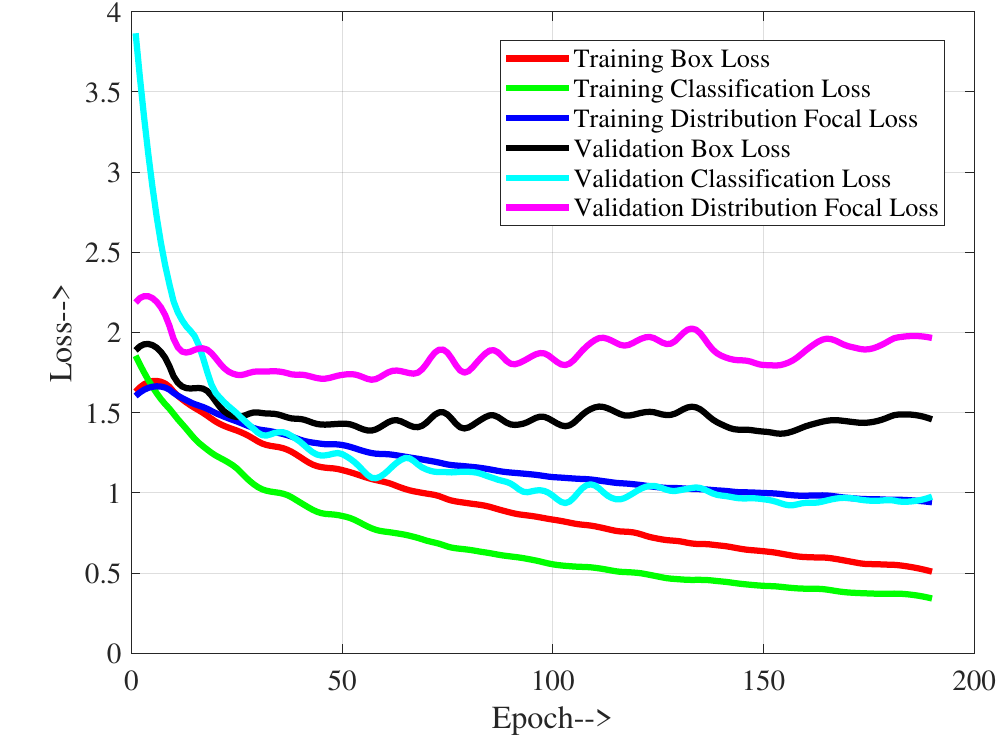}
    }\hfill
    \subfloat[]{
        \includegraphics[width=0.45\linewidth, trim=50 200 80 220, clip]{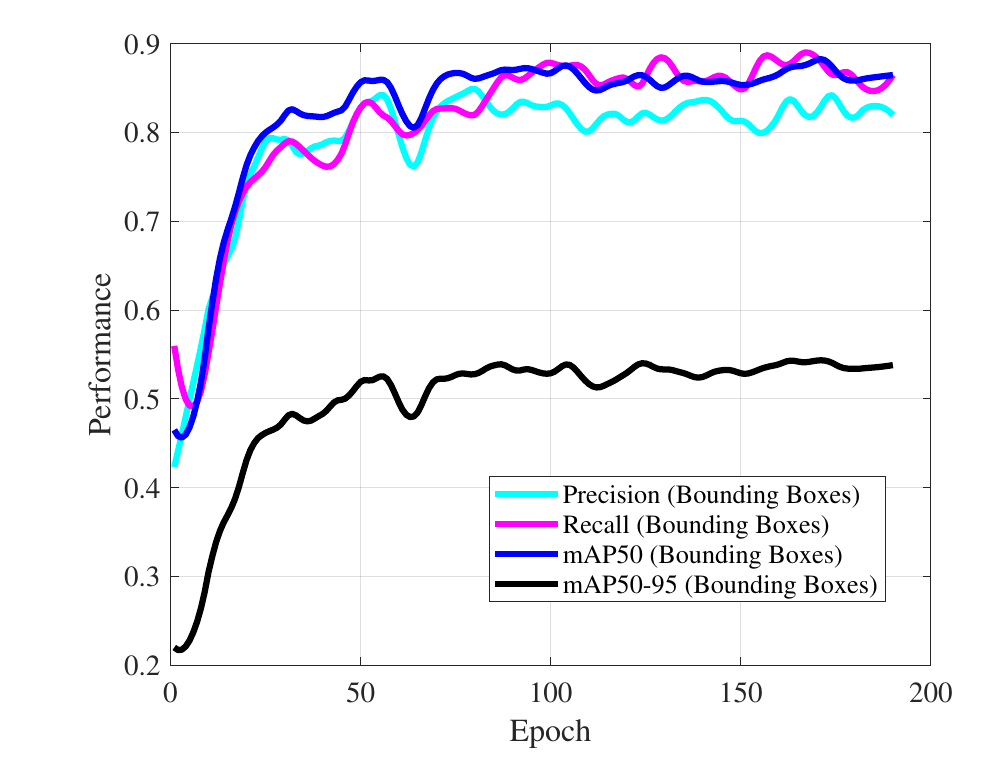}
    }
    \caption{Training behavior and qualitative detection performance of the proposed SOD module: a) Training losses of the SOD across epochs, b) Qualitative bounding-box predictions of the SOD.}
    \label{fig:combo}
\end{figure*}

Figure~\ref{fig:combo} illustrate the training dynamics and bounding-box predictions. On the USE50k test split, the SOD achieved an mAP@50 of~0.877 and mAP@50–95 of~0.599, with an inference latency of~6.2~ms per image, demonstrating real-time capability. To evaluate cross-dataset generalization, the SOD was tested on MPII, Market-1501, and Adience datasets (Table~\ref{table2}). These datasets contain fewer classes and more controlled environments than USE50k; hence the detector achieved higher mAP values, 0.905 (MPII), 0.933 (Market-1501), and 0.94 (Adience). The performance gap highlights the greater visual diversity, small-object density, and environmental complexity of the USE50k benchmark.


\begin{table}[h!]
\centering
\caption{Cross-Dataset Performance of the SOD}
\label{table2}
\begin{tabular}{p{1.6cm}|p{1cm}p{1.3cm}p{1cm}p{1.4cm}}
\hline
\textbf{Database} & \textbf{mAP50} & \textbf{mAP50-95} & \textbf{Classes} & \textbf{Latency (ms)}\\
\hline
MPII \cite{mpi} & 0.905 & 0.689 & 2 & 5.9\\
Market \cite{market1501} & 0.933 & 0.723 & 2 & 5.7\\
Adience \cite{Edience}   & 0.940 & 0.778 & 1 & 5.9\\
USE50k                 & 0.877 & 0.599 & 4 & 6.2\\
\hline
\end{tabular}
\end{table}

To further assess robustness, the proposed model was compared with several state-of-the-art detectors (TinyFace, EXTD, SSH, DRN, Face-MagNet, Face-RCNN, RetinaFace) on the USE50k dataset split into two subsets. Set-1 contains relatively simple scenes, while Set-2 represents challenging real-world conditions with low-resolution imagery, dense crowds, and small hazardous objects.

As summarized in Table~\ref{table3}, all competing detectors achieved high performance on Set-1 ($>$0.77 mAP), but exhibited performance drops on Set-2 due to its higher complexity. Region-based methods such as Face-RCNN showed stronger degradation. In contrast, the proposed SOD consistently outperformed or matched these SOTA approaches across all four classes (person, face, fire, weapon), demonstrating stronger resilience to occlusion, scale variation, and background clutter.

\begin{table*}[t]
\centering
\caption{Comparison of the Proposed SOD with State-of-the-Art Detectors (mAP).}
\label{table3}
\begin{tabular}{l|llrrrrrrrr}
\hline
\multirow{2}{*}{\textbf{Network}} & \multirow{2}{*}{\textbf{Backbone}} & \multirow{2}{*}{\textbf{Approach}} &
\multicolumn{2}{c}{\textbf{Person}} &
\multicolumn{2}{c}{\textbf{Face}} &
\multicolumn{2}{c}{\textbf{Fire}} &
\multicolumn{2}{c}{\textbf{Weapon}}\\
 & & &
\textbf{Set\_1} & \textbf{Set\_2} &
\textbf{Set\_1} & \textbf{Set\_2} &
\textbf{Set\_1} & \textbf{Set\_2} &
\textbf{Set\_1} & \textbf{Set\_2}\\
\hline
Tiny Face \cite{tiny}     & ResNet101 & Context reasoning & 0.78 & 0.75 & 0.77 & 0.69 & 0.73 & 0.66 & 0.78 & 0.74\\
EXTD \cite{EXTD}          & MobileNet & Feature fusion    & 0.774 & 0.729 & 0.765 & 0.658 & 0.723 & 0.644 & 0.778 & 0.739\\
SSH \cite{ssh}            & VGG16     & Context reasoning & 0.78 & 0.76 & 0.80 & 0.73 & 0.68 & 0.61 & 0.73 & 0.66\\
DRN \cite{DRN}            & ResNet50  & Anchor matching   & 0.79 & 0.76 & 0.79 & 0.74 & 0.73 & 0.66 & 0.77 & 0.73\\
Face-MagNet \cite{facemgt}& VGG16     & Context reasoning & 0.776 & 0.731 & 0.771 & 0.684 & 0.698 & 0.667 & 0.779 & 0.736\\
Face-RCNN \cite{facercnn} & VGG19     & Anchor matching   & 0.786 & 0.691 & 0.788 & 0.625 & 0.708 & 0.618 & 0.770 & 0.699\\
RetinaFace \cite{retina}  & RetinaNet & Context reasoning & 0.801 & 0.766 & 0.815 & 0.776 & 0.786 & 0.721 & 0.800 & 0.726\\
\textbf{Proposed}         & CSPDarknet53 & Feature fusion & \textbf{0.829} & \textbf{0.776} & \textbf{0.833} & \textbf{0.790} & \textbf{0.806} & \textbf{0.728} & \textbf{0.836} & \textbf{0.750}\\
\hline
\end{tabular}
\end{table*}

Overall, the comprehensive evaluation demonstrates that the proposed SOD delivers competitive accuracy, superior robustness under unconstrained conditions, and reliable real-time performance, making it well suited for downstream suspiciousness estimation in surveillance environments.

\subsection{Facial Expressions and Body Language Analysis}

\subsubsection{Training and Evaluation of the DCNNs}
The proposed DCNN architectures for facial expression (DCNN-I) and body language recognition (DCNN-II) were trained using minibatch sizes $\{16, 32, 64, 128, 256\}$, learning rates $\{0.1, 0.01, 0.03, 0.001, 0.003, 0.0001, 0.00001\}$, and multiple optimizers. The optimal configuration, batch size 128 and learning rate $10^{-4}$, showed stable convergence between 90-100 epochs, after which early stopping was applied.
Among SGDM, RMSProp, and Adam, the Adam optimizer yielded the best results, achieving mAP scores of 0.970 (emotion) and 0.744 (body language). Table~\ref{table4} summarizes the performance across optimizers.

\begin{table}[!t]
\centering
\caption{Performance of DCNNs using different optimizers}
\label{table4}
\begin{tabular}{l|rrrr}
\hline
\multirow{2}{*}{\textbf{Optimizer}} & \multicolumn{2}{c}{\textbf{DCNN-I}} & \multicolumn{2}{c}{\textbf{DCNN-II}}\\
\cline{2-5}
 & \textbf{mAP} & \textbf{Recall} & \textbf{mAP} & \textbf{Recall} \\
\hline
SGDM    & 0.964 & 0.930 & 0.669 & 0.645\\
RMSProp & 0.966 & 0.945 & 0.723 & 0.711\\
Adam    & 0.970 & 0.956 & 0.744 & 0.691\\
\hline
\end{tabular}
\end{table}

\subsubsection{Comparative Analysis with State-of-the-Art}
DCNN-I was benchmarked on standard FER datasets using uniform training settings. It achieved accuracies of 97\%, 97.63\%, 72.20\%, 95.60\%, and 64.50\% on LNMIIT \cite{kd_mtap}, CK+ \cite{ck}, FER2013 \cite{fer2013}, FER20E \cite{iitd-fer}, and AffectNet \cite{affectnet}, respectively. Lower performance on FER2013 and AffectNet is expected due to high intra-class diversity, label noise, and subtle/compound emotion categories.

\begin{table}[h]
\centering
\caption{Comparison with State-of-the-Art Methods (mAP)}
\label{Table:table6}
\begin{tabular}{l|ll}
\hline
\textbf{Dataset} & \textbf{Method} & \textbf{mAP} \\
\hline
\multirow{8}{*}{AffectNet}
 & ResNet \cite{Resenet_71}            & 0.607 \\
 & InceptionV2 \cite{Ince_72}          & 0.599 \\
 & MobileNet \cite{Mobile_74}          & 0.617 \\
 & VGG16 \cite{VGG16_73}               & 0.612 \\
 & Xception \cite{Xcep_75}             & 0.595 \\
 & Inception-ResNet \cite{Incep_76}    & 0.595 \\
 & BReG-Net \cite{BReG}                & 0.638 \\
 & \textbf{Ours}                       & \textbf{0.648} \\
\hline
\multirow{10}{*}{FER2013}
 & Going Deeper \cite{Going_77}        & 0.664 \\
 & FER2013 Winner \cite{FER2013_win}   & 0.712 \\
 & MDNL \cite{MDNL_79}                 & 0.720 \\
 & Adaptive Weighting \cite{Adaptive_80} & 0.726 \\
 & H-DCNN \cite{H_DCNN_81}             & 0.727 \\
 & Multi-scale CNNs \cite{multi_82}    & 0.728 \\
 & Custom CNN \cite{CNN}               & 0.664 \\
 & ZFER-FCNN \cite{shahzad2023}        & 0.650 \\
 & SML \cite{sml}                      & 0.728 \\
 & \textbf{Ours}                       & \textbf{0.730} \\
\hline
\multirow{6}{*}{CK+}
 & Custom CNN \cite{CNN}               & 0.932 \\
 & CNN+OpLoss \cite{OpLoss}            & 0.955 \\
 & FCNN \cite{fcnn}                    & 0.844 \\
 & SBN-CNN \cite{r3}                    & 0.968 \\
 & ZFER-FCNN \cite{shahzad2023}        & 0.970 \\
 & \textbf{Ours}                       & \textbf{0.985} \\
\hline
\multirow{3}{*}{LNMIIT}
 & KNN+fusion \cite{kd_mtap}           & 0.962 \\
 & KNN+distance \cite{Kd_ISCON}        & 0.960 \\
 & CNN+distance \cite{Kd_spices}       & 0.965 \\
 & \textbf{Ours}                       & \textbf{0.978} \\
\hline
\textbf{FER20E} & \textbf{Ours}        & \textbf{0.920} \\
\hline
\end{tabular}
\end{table}

Across these benchmarks, the proposed model performs competitively on challenging datasets (AffectNet, FER2013) and surpasses existing approaches on controlled datasets (CK+, LNMIIT, FER20E), reflecting its strong generalization across varied expression intensities and acquisition conditions.

For body language analysis, DCNN-II was trained using the same protocol and evaluated on USE50k. It achieved an mAP of 0.85 across the two categories (normal vs. abnormal behavior), demonstrating the effectiveness of lightweight landmark-guided representations in capturing subtle behavioral cues under pose variations, occlusion, and clutter.

\subsection{Experimental Analysis of Feature Aggregation and Discriminator}

The final stage of the \textit{DeepUSEvision} framework employs the Feature Aggregation and Discriminator (FAD) module to integrate multimodal cues extracted from the SOD, facial expression, and body language analyzers for unified suspiciousness estimation. A total of 65,000 samples from the USE50k dataset were considered for this experiment after discarding blurred and heavily distorted instances. Each sample is annotated with a ground-truth risk score in the range of 0–10, representing the perceived suspiciousness level under real-world surveillance constraints.

The FAD network was trained for 50 epochs using a batch size of 16, a learning rate of $1 \times 10^{-4}$, and the Adam optimizer. Early stopping was enabled with a patience of 8 epochs to mitigate overfitting. The regression-based performance was quantitatively evaluated using four complementary metrics: Mean Absolute Error (MAE), Root Mean Square Error (RMSE), Coefficient of Determination ($R^2$), and Mean Absolute Percentage Error (MAPE). The obtained results are summarized in Table~\ref{tab:results_fad}.  

\begin{table}[hbt]
\centering
\caption{Performance Metrics of Feature Aggregation and Discriminator}
\label{tab:results_fad}
\begin{tabular}{lcc}
\hline
\textbf{Metric} & \textbf{Training} & \textbf{Validation} \\ 
\hline
MAE  & 0.412 & 0.466 \\
RMSE & 0.613 & 0.684 \\
$R^2$ & 0.941 & 0.935 \\
MAPE (\%) & 7.41 & 8.93 \\
\hline
\end{tabular}
\end{table}

\begin{figure}[h]
    \centering
    \includegraphics[width=1\linewidth, trim=10 10 10 30, clip]{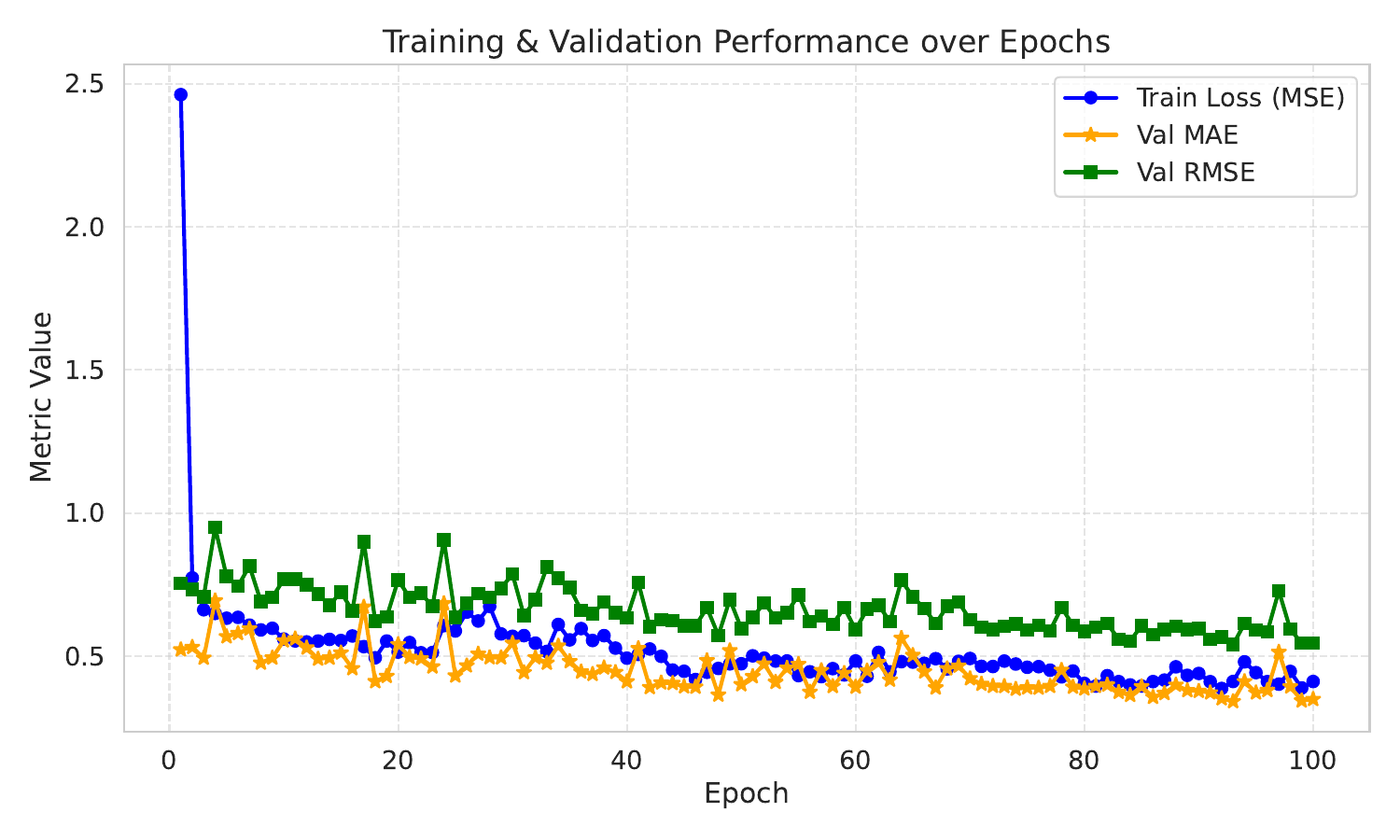}
    \caption{Training Dynamics of the Proposed Transformer-Based Discriminator}
    \label{fig:Training_Dis}
\end{figure}
\begin{figure}[h]
    \centering
    \includegraphics[width=1\linewidth, trim=10 10 10 30, clip]{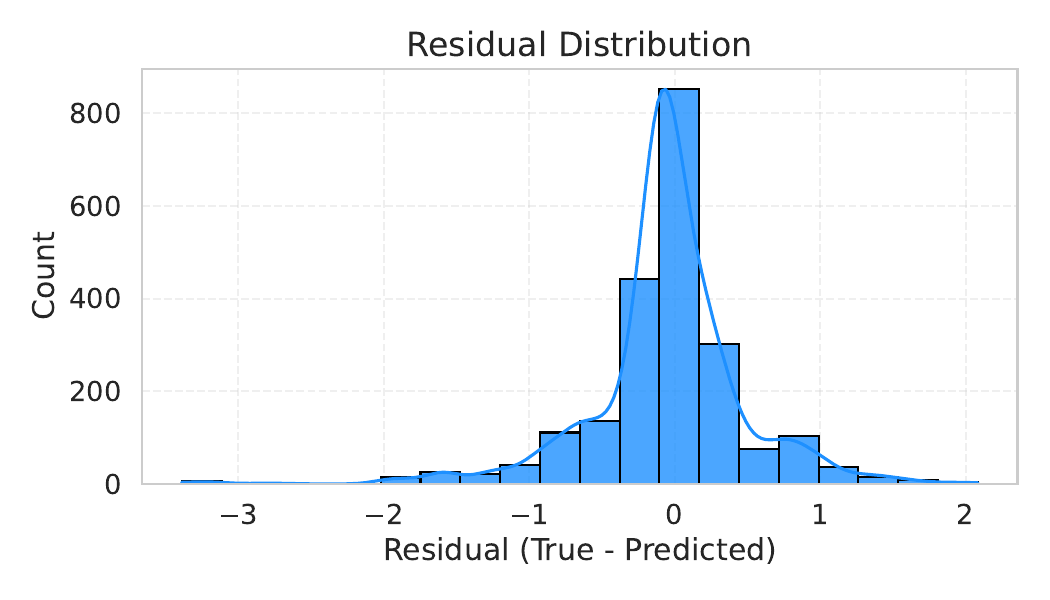}
    \caption{Statistical Residual Distribution of Risk Score Prediction}
    \label{fig:Residual}
\end{figure}

The low MAE and RMSE values confirm that the FAD module achieves highly accurate score predictions with minimal deviation from the ground truth. Furthermore, an $R^2$ score of approximately 0.94 verifies a strong positive correlation between predicted and true risk levels, while a MAPE below 10\% indicates stable performance across varying object density, occlusion, illumination, and behavioral uncertainty conditions.
The training curve exhibits a consistent downward trend in the training loss (MSE), confirming stable convergence of the Transformer-based discriminator, as illustrated in Fig.~\ref{fig:Training_Dis}. The validation MAE and RMSE follow a similar trajectory, indicating that the model maintains strong generalization capability without signs of overfitting. The narrow and persistent gap between training and validation curves further validates the stability of the proposed fusion framework, reflecting disciplined learning behavior and effective regularization. Moreover, the predicted risk scores, plotted against ground truth in Fig.~\ref{fig:Residual}, lie densely along the identity line with minimal dispersion, confirming high prediction fidelity.

\begin{figure}[hbt]
    \centering
    \includegraphics[width=1\linewidth, trim=10 10 10 30, clip]{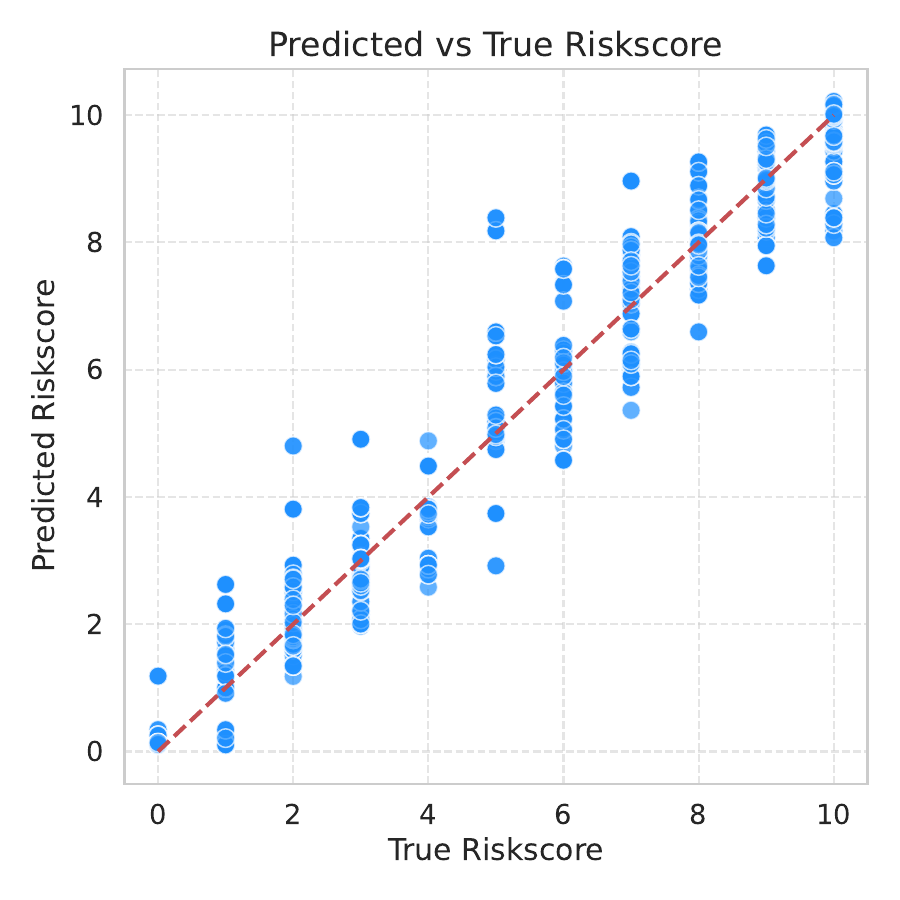}
    \caption{Ground Truth vs. Predicted Suspiciousness Scores}
    \label{fig:PridVStrue}
\end{figure}

No systematic bias toward overestimation or underestimation is observed across low-, medium-, and high-risk regions, demonstrating reliable behavior-critical ranking performance under real-world surveillance conditions. The residual error distribution in Fig.~\ref{fig:PridVStrue} is sharply centered near zero, forming a symmetric Gaussian-like profile with no heavy tails. This signifies that the discriminator does not exhibit unstable performance spikes or critical failure zones, a crucial reliability trait for deployment in security-sensitive settings.

To further assess robustness, controlled perturbation experiments were conducted. Gaussian noise ($\sigma = 0.05$) was injected into the multimodal embeddings, resulting in less than 3\% degradation in overall accuracy, confirming strong noise tolerance. Excluding any single modality (e.g., facial or body cues) caused only a marginal drop of 1.8\% in $R^2$, demonstrating the resilience and redundancy built into the fusion strategy. Additionally, temporal misalignment of $\pm2$ frames introduced only a negligible shift in prediction error ($\Delta \text{MAE} \approx 0.02$). These observations collectively validate the robustness and generalization strength of the proposed transformer-based aggregation mechanism under real-world operational disturbances.

\begin{figure}[b]
    \centering
    \includegraphics[width=1\linewidth,trim=250 100 230 110,clip]{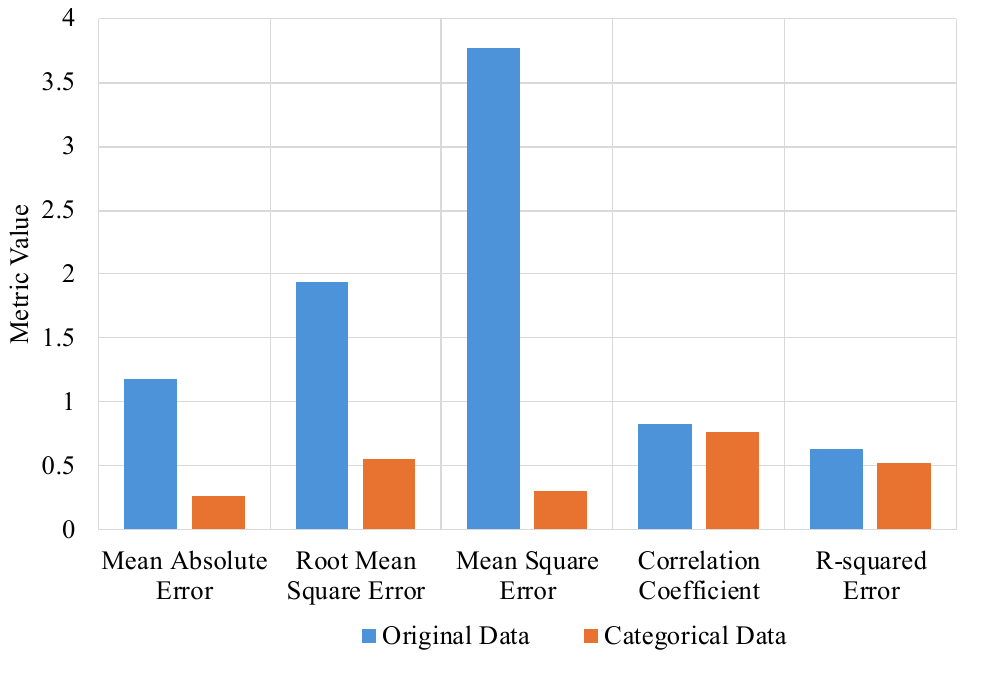}
    \caption{Discriminator Performance: Various performance metrics.}
    \label{fig:attention_heatmap}
\end{figure}

To further evaluate the reliability of the risk estimation, two complementary assessment strategies were employed: a) continuous regression using the original risk scale (0–10), and b) discretized risk classification using a quantized scale.  
In the latter, the predicted risk scores were mapped into three semantic classes, \textit{Low} (0–3), \textit{Medium} (4–6), and \textit{High} (7–10).  
This quantization significantly reduced decision complexity while preserving discriminative intent, resulting in improved stability and operational interpretability. The impact of this categorization strategy is visually reflected in Fig.~\ref{fig:attention_heatmap}.
\subsection{Explainability and Interpretability Analysis}
To validate the reliability and transparency of the proposed Discriminator Network, a comprehensive explainability analysis was performed, progressing from linear dependencies to higher-order nonlinear interactions among visual cues. This multi-level interpretation ensures that the fusion model not only performs accurately but also provides clear justification for its decisions, a crucial requirement in forensic and surveillance domains.

\subsubsection{Correlation and Statistical Dependencies}
\begin{figure}[t!]
    \centering
    \includegraphics[width=1\linewidth]{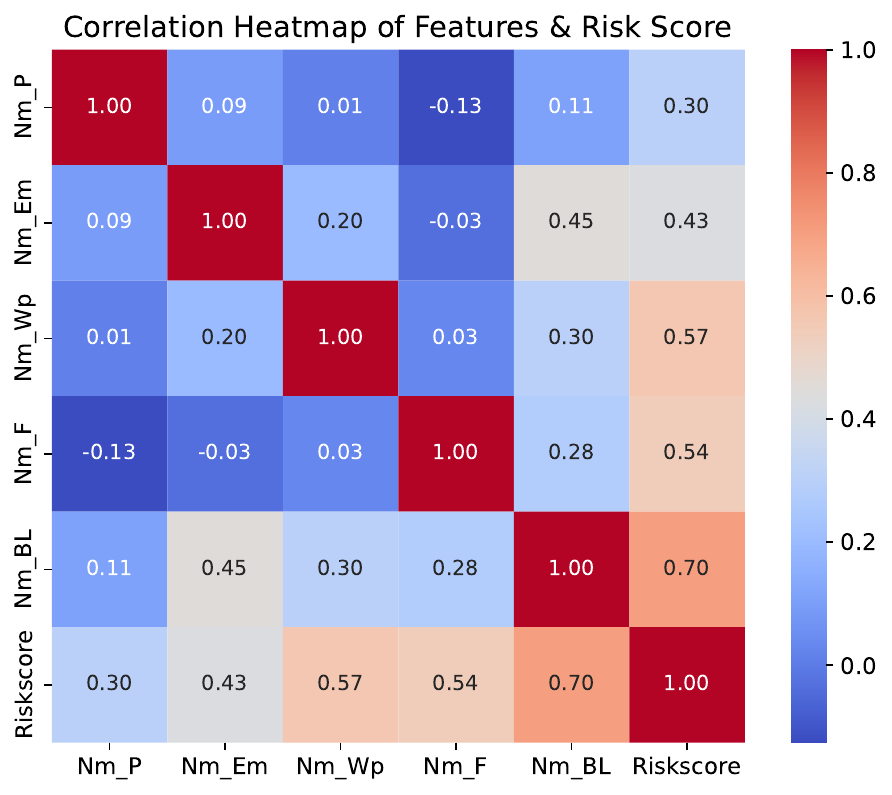}
    \caption{Correlation heatmap showing pairwise linear dependencies between visual cues and the risk score. Strong associations with Nm\_Wp, Nm\_BL, and Nm\_F confirm their dominant statistical influence.}
    \label{fig:correlation}
\end{figure}

We first examined the linear interdependence among input cues and the predicted risk score using Pearson correlation analysis. As shown in Fig.~\ref{fig:correlation}, the correlation heatmap reveals distinct behavioral groupings: Nm\_Wp (weapon presence), Nm\_BL (abnormal body language),  and Nm\_F (fire flags) exhibit strong positive correlations with the risk score, while Nm\_P (number of persons) and Nm\_Em (facial expression) show moderate correlations, indicating that contextual crowding and behavioral anomalies have indirect but meaningful influence on suspiciousness escalation. The mild inter-feature correlations suggest that the cues contribute complementary information rather than redundant evidence.

\subsubsection{Global Feature Contribution Analysis}
To quantify global feature importance beyond linear correlations, three regression models, Linear Regression, Random Forest, and SHAP, were compared. Table~\ref{tab:feature_importance} presents the relative weight contributions for each cue. Consistently across all models, Nm\_Wp and Nm\_F emerge as the strongest predictors of the risk score, confirming their pivotal roles in threat inference. Interestingly, Random Forest and SHAP analyses emphasize Nm\_BL as highly significant, capturing its nonlinearly coupled behavior with other cues. 

\begin{table}[h!]
\centering
\caption{Feature importance comparison using Linear Regression, Random Forest, and SHAP.}
\begin{tabular}{|c|c|c|c|}
\hline
\textbf{Feature} & \textbf{Linear (\%)} & \textbf{Random Forest (\%)} & \textbf{SHAP (\%)} \\
\hline
Nm\_P   & 22.72 & 14.27 & 16.98 \\
Nm\_Em  & 13.53 &  4.97 & 11.08 \\
Nm\_Wp  & 22.16 & 15.63 & 23.91 \\
Nm\_F   & 24.81 & 15.10 & 25.46 \\
Nm\_BL  & 16.78 & 50.03 & 22.57 \\
\hline
\end{tabular}
\label{tab:feature_importance}
\end{table}

\begin{figure}[h!]
    \centering
    \includegraphics[width=1\linewidth,trim=210 120 150 70,clip]{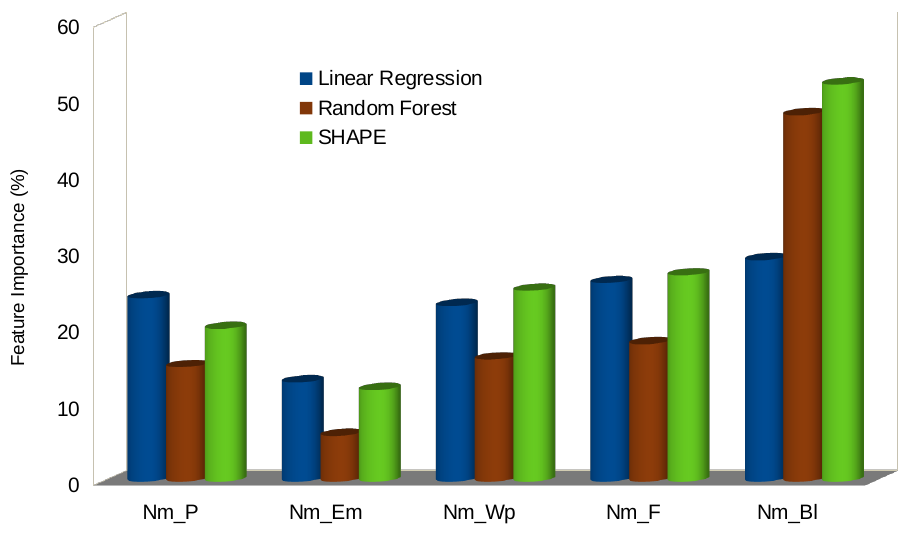}
    \caption{Global feature importance obtained using SHAP values.}
    \label{fig:feature_importance}
\end{figure}

The SHAP-based global feature ranking (Fig.~\ref{fig:feature_importance}) provides model-agnostic interpretability. It reveals that while Nm\_Wp and Nm\_F dominate overall suspiciousness, the Nm\_BL and Nm\_Em cues play a secondary yet contextually amplified role when co-occurring with primary threat indicators.

\begin{figure}[h]
    \centering
    \includegraphics[width=\linewidth]{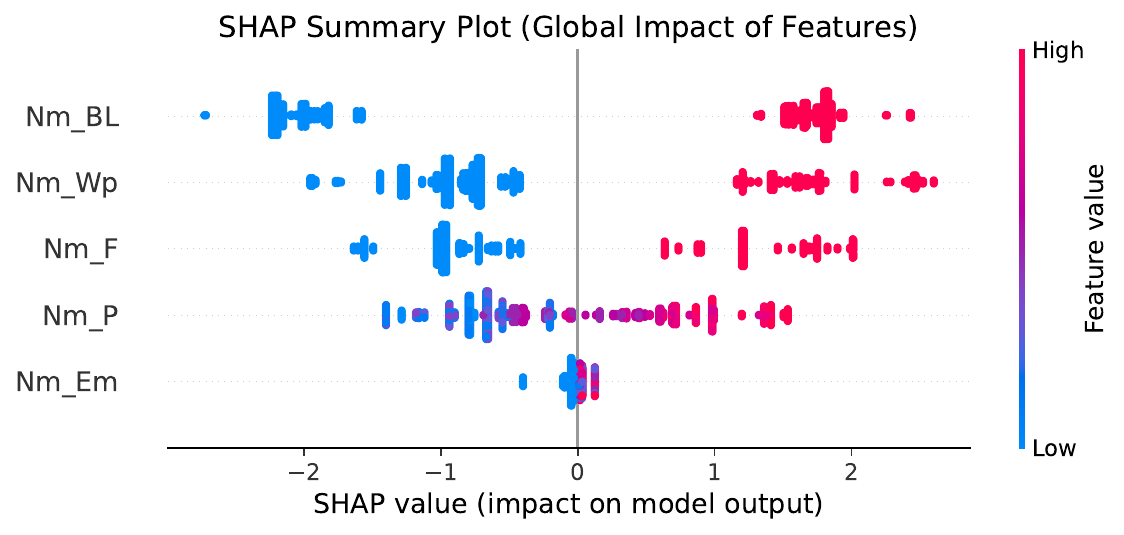}
    \caption{SHAP dependence plots showing pairwise feature interactions. Strong nonlinear amplification is observed for Nm\_Wp with Nm\_Em and Nm\_BL combinations.}
    \label{fig:shap_dependence}
\end{figure}

\begin{figure*}[h]
    \centering
    \subfloat[]{
        \includegraphics[width=0.3\linewidth, trim=10 10 10 30,clip]{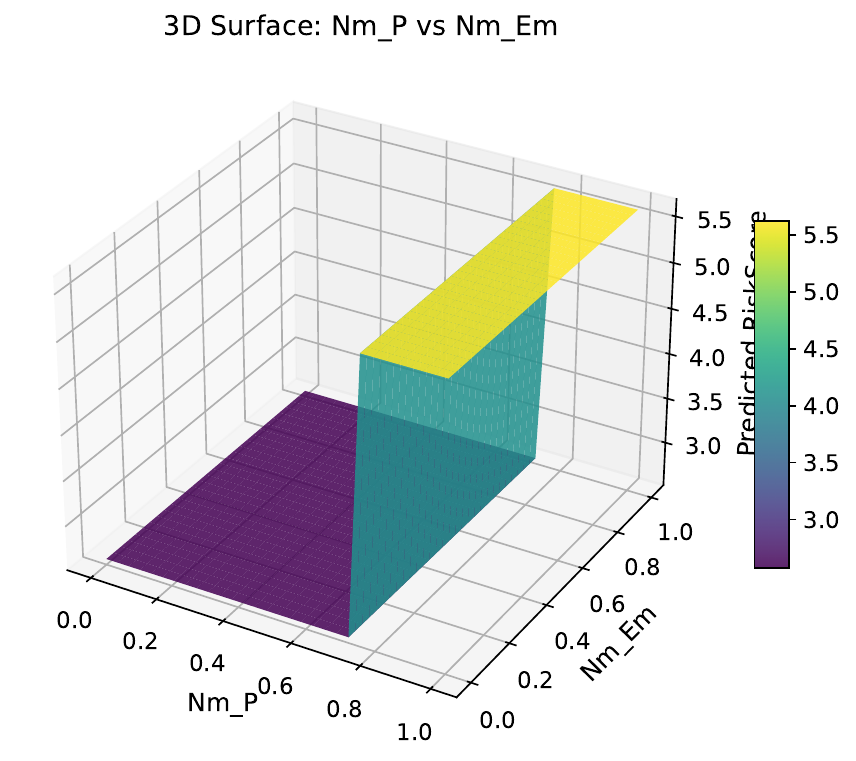}
    }\hfill
    \subfloat[]{
        \includegraphics[width=0.3\linewidth, trim=10 10 10 30,clip]{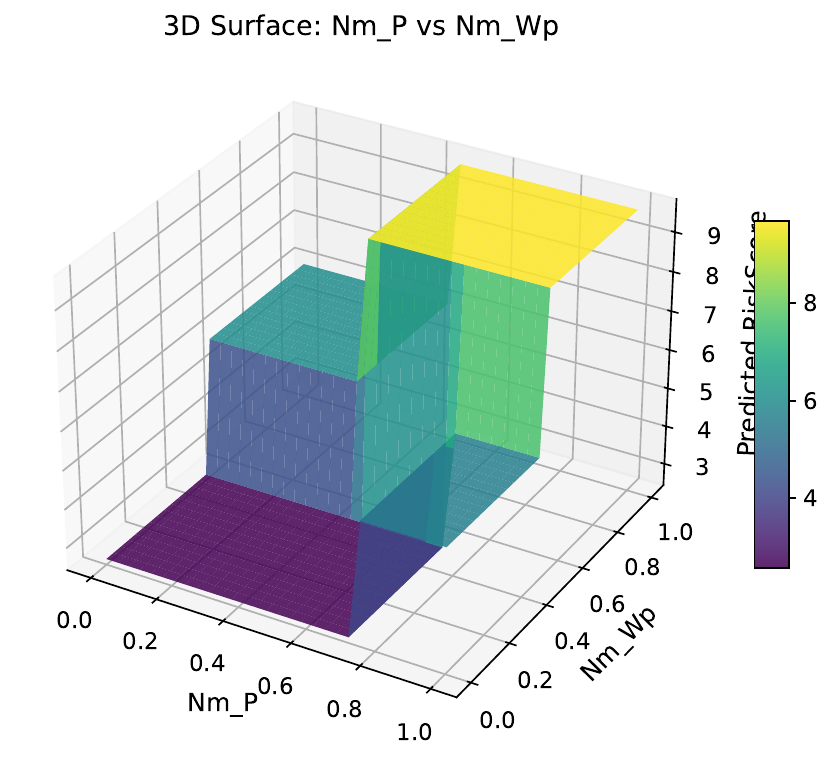}
    }\hfill
    \subfloat[]{
        \includegraphics[width=0.3\linewidth, trim=10 10 10 30,clip]{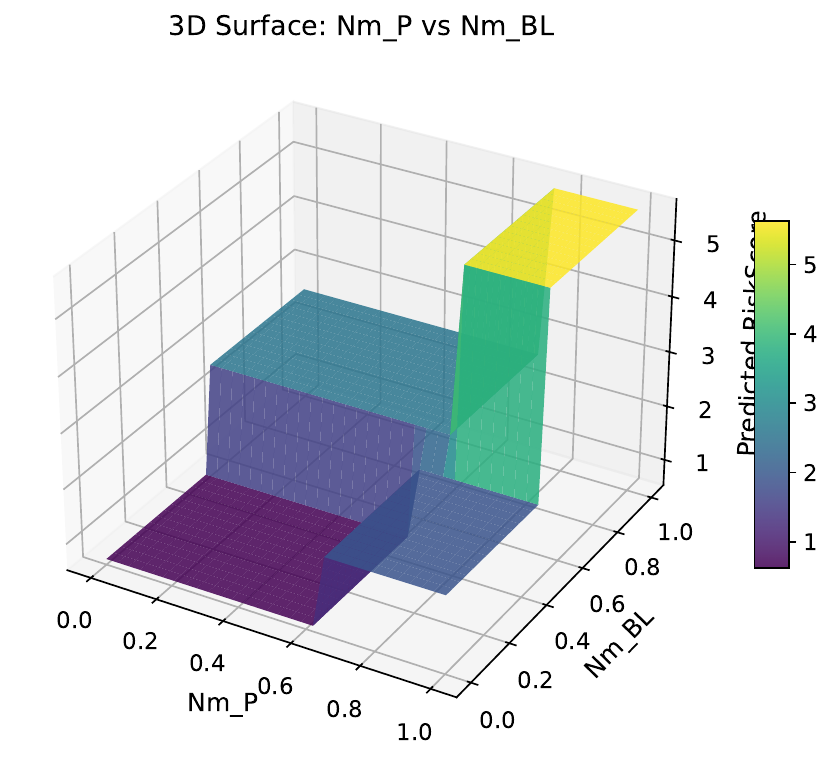}
    }\\[1ex]
    \subfloat[]{
        \includegraphics[width=0.3\linewidth, trim=10 10 10 30,clip]{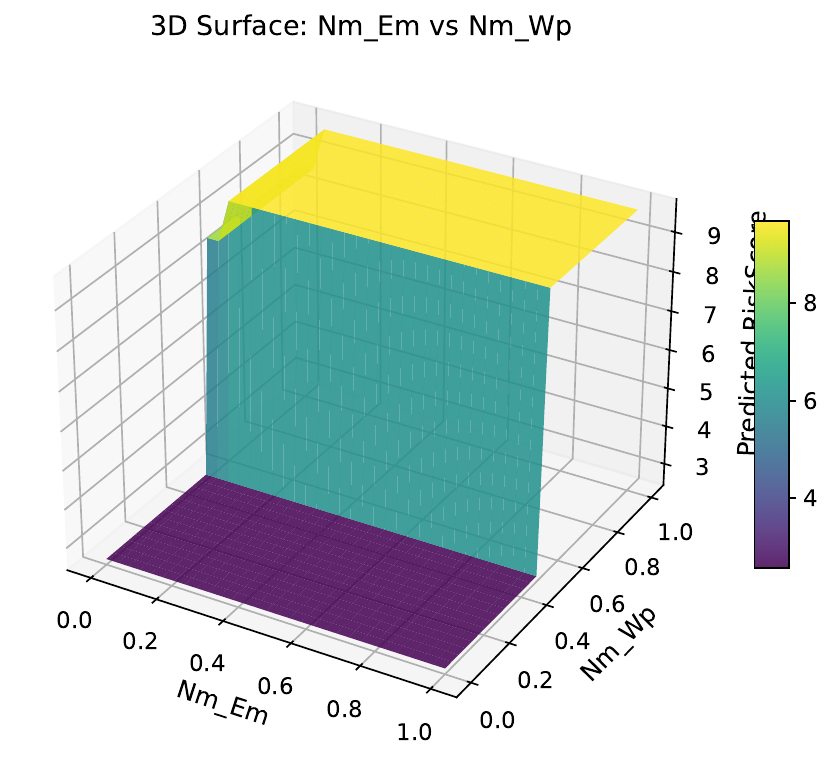}
    }\hfill
    \subfloat[]{
        \includegraphics[width=0.3\linewidth, trim=10 10 10 30,clip]{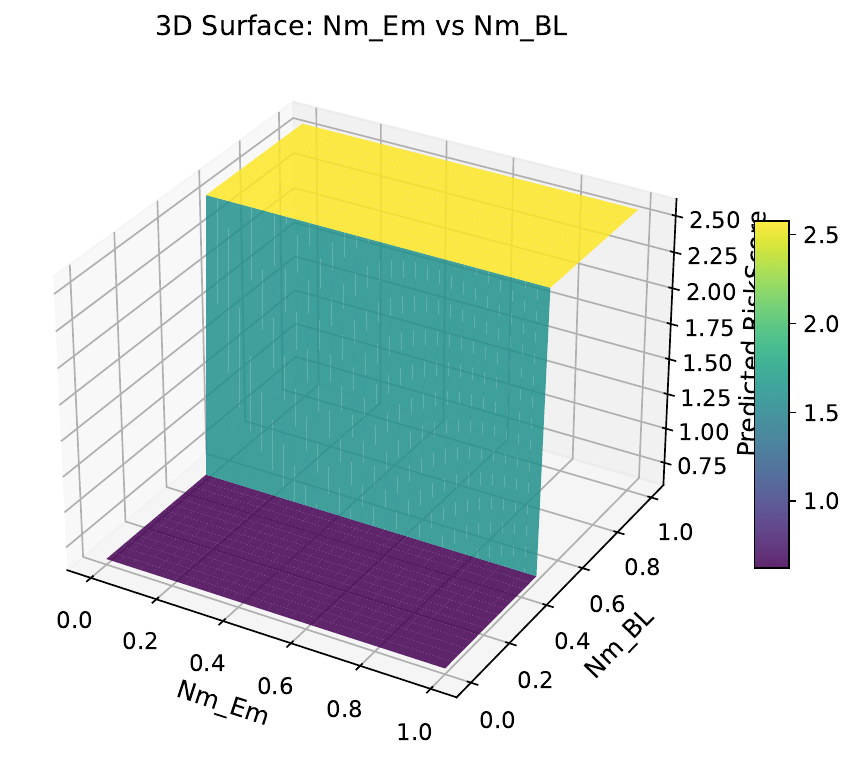}
    }\hfill
    \subfloat[]{
        \includegraphics[width=0.3\linewidth, trim=10 10 10 30,clip]{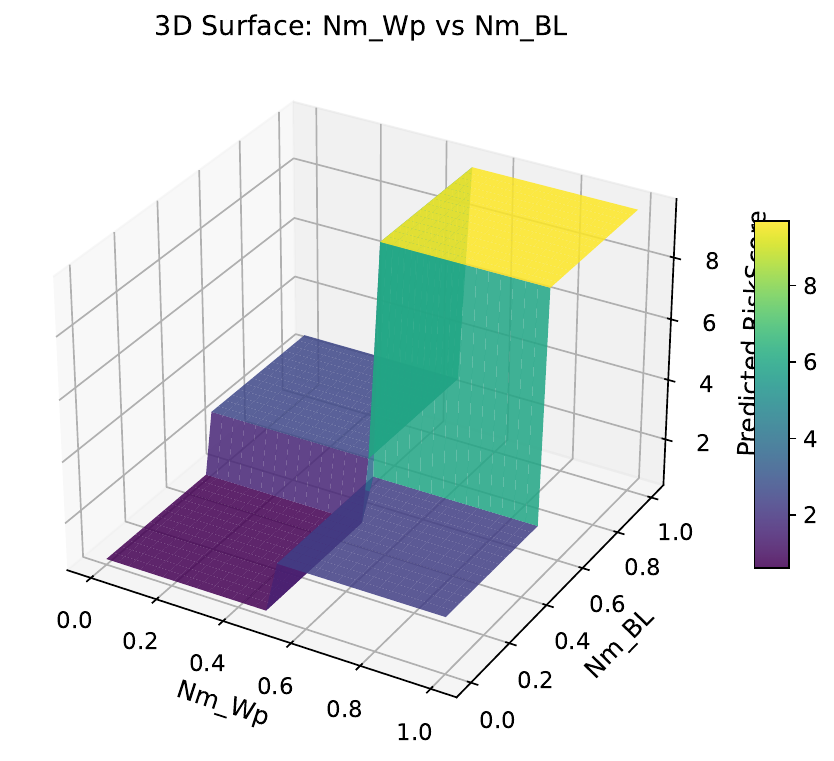}
    }
    \caption{3D surface visualization of learned risk manifolds across feature pairs. Steeper gradients reflect high-risk interaction zones dominated by Nm\_Wp and Nm\_Em combinations: a) Person–Emotion Interaction, b) Person–Weapon Interaction, c) Person–Body Language Interaction, d) Emotion–Weapon Interaction, e) Emotion–Body Language, f) Weapon–Body Language}
    \label{fig:3d_surface}
\end{figure*}

\subsubsection{Pairwise and Higher-Order Interactions}
Beyond global influence, SHAP interaction values were computed to explore synergistic relationships among cues. Fig.~\ref{fig:shap_dependence} presents the SHAP dependence plots, which illustrate how the presence of one cue modulates the impact of another. Notably, the combination of Nm\_Wp (weapons) with Nm\_Em (suspicious emotions) produces a sharp nonlinear rise in the predicted risk, implying that emotional distress in proximity to a weapon strongly escalates system suspicion. Similarly, coupling between Nm\_Wp and Nm\_BL (abnormal body language) yields high joint activation, highlighting behavioral–object synergy in threat perception.

\subsubsection{Nonlinear Manifold Visualization through 3D Surface Analysis}
To capture complex non-additive effects, the learned risk manifold was visualized via 3D surface plots (Fig.~\ref{fig:3d_surface}). These surfaces model the response of the Discriminator to varying combinations of two cues while holding others constant. Sharp curvature regions correspond to high-sensitivity zones where small input perturbations trigger large risk score variations. The plots clearly show that interactions involving Nm\_Wp act as dominant attractors on the manifold, especially when coupled with Nm\_Em or Nm\_BL. Such nonlinear topography confirms that the proposed Transformer fusion effectively encodes higher-order dependencies critical for situational awareness.

\subsubsection{Feature Synergy Network Representation}
\begin{figure}[t!]
    \centering
    \includegraphics[width=0.8\linewidth, trim = 0 0 0 30, clip]{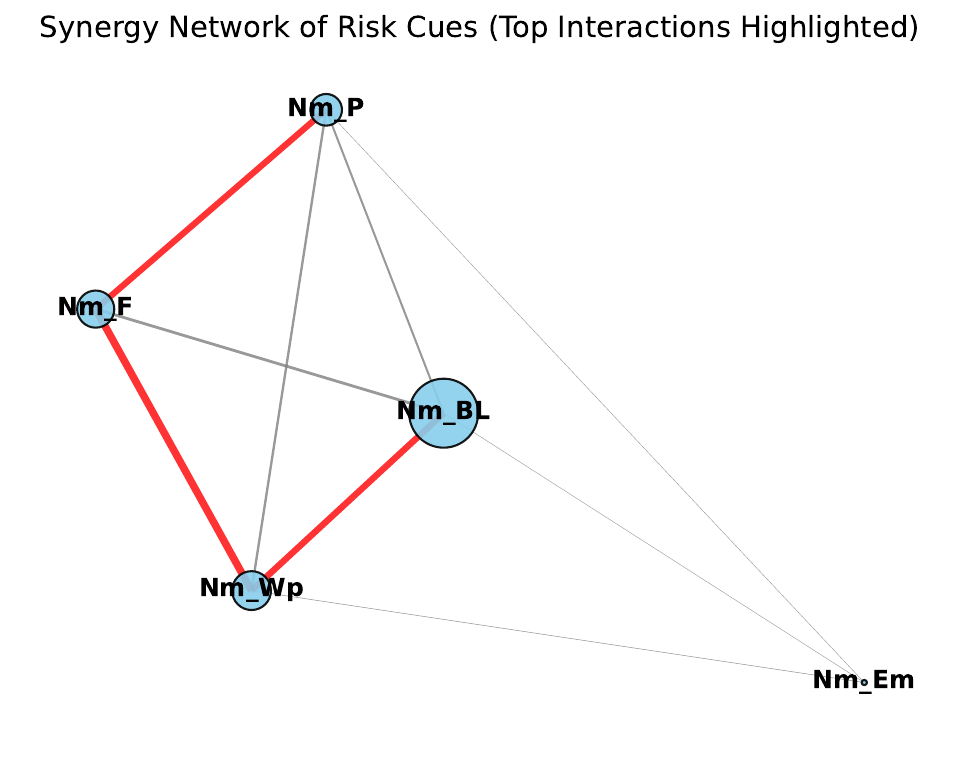}
    \caption{Feature synergy network summarizing pairwise cue interactions. Node size $\propto$ global importance; edge width $\propto$ SHAP interaction strength. Red edges denote top-3 synergies.}
    \label{fig:synergy_network}
\end{figure}

To consolidate the multi-level findings, a feature synergy network was constructed (Fig.~\ref{fig:synergy_network}). Nodes represent features (size proportional to global SHAP importance), while edges encode interaction strength derived from SHAP interaction values. The top-3 strongest couplings, listed in Table~\ref{tab:top3_interactions}, show that Nm\_Wp – Nm\_Em, Nm\_Wp – Nm\_BL, and Nm\_Em – Nm\_BL form the dominant subgraph. The resulting topology reveals a star-like structure centered on Nm\_Wp, reflecting its catalytic role in suspiciousness formation.

\begin{table}[h!]
    \centering
    \caption{Top-3 strongest SHAP-based feature interactions in the Discriminator Network.}
    \begin{tabular}{|c|c|c|}
        \hline
        \textbf{Rank} & \textbf{Feature Pair} & \textbf{Interaction Strength} \\
        \hline
        1 & Nm\_Wp \& Nm\_Em & 0.2187 \\
        2 & Nm\_Wp \& Nm\_BL & 0.1902 \\
        3 & Nm\_Em \& Nm\_BL & 0.1654 \\
        \hline
    \end{tabular}
    
    \label{tab:top3_interactions}
\end{table}


The integrated explainability analysis reveals that suspiciousness emerges from nonlinear multimodal interactions rather than isolated cues. Weapon-related features (\textit{Nm\_Wp}) exhibit the highest global influence, serving as dominant predictors of risk, while emotional cues (\textit{Nm\_Em}) amplify contextual threat perception when co-occurring with weapon or abnormal behavior indicators. The body language component (\textit{Nm\_BL}) reinforces suspiciousness, particularly when aligned with hostile emotional states or object-based anomalies. Crowd-related features (\textit{Nm\_P}) contribute indirectly by modulating perceived threat intensity through population density, whereas fire cues (\textit{Nm\_F}) introduce discrete yet localized risk spikes. Overall, the Discriminator Network effectively models these synergistic dependencies, ensuring both predictive precision and interpretability, crucial for reliable deployment in real-world surveillance environments.
\subsection{Inference-Time and Computational Efficiency Analysis}
To evaluate the real-time feasibility of the proposed surveillance framework, we performed a detailed inference-time and computational efficiency analysis across all major processing stages. The full pipeline consists of four key components: (i) the Suspicious Object Detector (SOD) based on YOLOv12, (ii) two lightweight convolutional networks, DCNN-I for facial emotion recognition and DCNN-II for body-language classification, (iii) the Transformer-based Discriminator Network for contextual suspiciousness estimation, and (iv) a feature aggregation module for multimodal fusion.

Under the configuration of the proposed setup, the YOLOv12-based SOD achieved an average throughput of approximately 28.6~FPS, corresponding to a per-frame latency of 34.9~ms. This stage constitutes the primary computational bottleneck due to its multi-scale feature fusion layers, dense anchor-free prediction heads, and the high-resolution inputs typically encountered in surveillance videos. The lightweight DCNN-I and DCNN-II networks operate with substantially lower computational overhead. DCNN-I achieves a throughput of 42.3~FPS (23.6~ms/frame), while DCNN-II operates at 39.5~FPS (25.3~ms/frame). Their efficient depthwise-separable convolutional design enables rapid extraction of facial expression and body-language cues without compromising accuracy. The Transformer-based Discriminator exhibits the lowest latency among all modules, with a runtime of 19.5~ms per frame (51.2~FPS). Owing to its compact token dimensionality and shallow attention blocks, the contextual interaction modeling of multimodal cues introduces minimal overhead relative to the detection stage.

Integrating all stages into a unified execution pipeline yields an end-to-end throughput of approximately 16.7~FPS, equivalent to a total latency of 59.8~ms per frame. This satisfies real-time operational requirements for intelligent surveillance, where frame rates of 10--15~FPS are typically considered adequate for automated threat detection. The results confirm that the proposed framework achieves real-time performance while maintaining high-level contextual reasoning and explainability.

\section{Conclusion and Future Work}

This work presents a comprehensive and explainable framework for visual suspiciousness estimation, integrating advances from computer vision, image processing, and deep learning. The proposed system comprises three major components: a Suspicious Object Detector (SOD), a Facial Expression Classifier (DCNN-I), and a Body-Language Analyzer (DCNN-II). Each module was independently trained, optimized, and benchmarked against state-of-the-art (SOTA) models. The SOD demonstrated a 1–2\% gain in mean Average Precision (mAP) over comparable detectors, while DCNN-I achieved a 1–5\% improvement in mAP across multiple benchmark datasets, including AffectNet, FER2013, CK+, LNMIIT, and FER20E. Similarly, DCNN-II achieved an mAP of 0.85 under an identical experimental setup, reinforcing the system’s robustness and modular consistency. The end-to-end system was evaluated using standard regression and classification metrics such as MAE, MSE, RMSE, correlation coefficient, and R-squared score. Two complementary assessment schemes were employed: a) continuous risk prediction on a 0–10 scale and b) quantized risk categorization into low (0–3), medium (4–6), and high (7–10) levels. The categorized approach exhibited higher accuracy and lower complexity, highlighting its suitability for practical deployment.
Furthermore, the introduction of the USE50k dataset establishes a new benchmark for multimodal suspiciousness estimation, validated through comparative analyses with existing datasets and models. The explainability study reveals that suspiciousness emerges as a nonlinear interaction among object, emotion, and body cues rather than as an additive effect of isolated features.

The future work will explore end-to-end fine-tuning across all modules through a unified transformer-fusion strategy. This will enable full gradient backpropagation and improve real-time adaptability without compromising modular interpretability.

\section*{Acknowledgments}
We acknowledge the Department of Electrical Engineering, Indian Institute of Technology Delhi, and CSIR Fourth Paradigm Institute Bengaluru, for their invaluable guidance and computational resources.

\section*{Data Availability Statement}
The data can be made available from the corresponding author upon reasonable request for academic and non-commercial research purposes.

\bibliographystyle{IEEEtran}
\bibliography{refs}
\vfill
\end{document}